\documentclass[conference]{IEEEtran}
\IEEEoverridecommandlockouts
\usepackage{cite}
\usepackage{amsmath,amssymb,amsfonts}
\usepackage{algorithmic}
\usepackage{graphicx}
\usepackage{textcomp}
\usepackage{xcolor}
\usepackage{gensymb}
\usepackage{subfigure}
\usepackage{orcidlink}
\def\BibTeX{{\rm B\kern-.05em{\sc i\kern-.025em b}\kern-.08em
    T\kern-.1667em\lower.7ex\hbox{E}\kern-.125emX}}
\begin{document}

\title{Physics-Informed Machine Learning for Pouch Cell Temperature Estimation\\
}

\author{\IEEEauthorblockN{Zheng Liu}
\IEEEauthorblockA{\textit{Department of Industrial and Manufacturing Systems Engineering} \\
\textit{University of Michigan-Dearborn}\\
Dearborn, USA \\
\orcidlink{0000-0003-4869-8893}\url{https://orcid.org/0000-0003-4869-8893}}

}

\maketitle

\begin{abstract}
Accurate temperature estimation of pouch cells with indirect liquid cooling is essential for optimizing battery thermal management systems for transportation electrification. However, it is challenging due to the computational expense of finite element simulations and the limitations of data-driven models. This paper presents a physics-informed machine learning (PIML) framework for the efficient and reliable estimation of steady-state temperature profiles. The PIML approach integrates the governing heat transfer equations directly into the neural network’s loss function, enabling high-fidelity predictions with significantly faster convergence than purely data-driven methods. The framework is evaluated on a dataset of varying cooling channel geometries. Results demonstrate that the PIML model converges more rapidly and achieves markedly higher accuracy, with a 49.1\% reduction in mean squared error over the data-driven model. Validation against independent test cases further confirms its superior performance, particularly in regions away from the cooling channels. These findings underscore the potential of PIML for surrogate modeling and design optimization in battery systems.
\end{abstract}


\section{Introduction}

The transition towards transportation electrification has amplified the demand for advanced energy storage systems \cite{liu2023life}. Batteries, as the core component of electric vehicles (EVs), are undergoing relentless evolution to improve energy density, operational safety, and cost efficiency \cite{liu2024machine}. Among available form factors of the batteries, pouch cells have progressively distinguished themselves as a promising candidate, thanks to their light weight, high energy density, design flexibility, and low manufacturing cost \cite{kim2007pouch}. Pouch cells, also a good fit for solid-state batteries, attract more attention from the frontier of battery research \cite{wang2024designing}.
A fundamental challenge for pouch cells is managing their thermal behavior, as temperature critically influences performance, cycle life, and safety \cite{belt2005effect}. The temperature must be maintained within the optical range to avoid reduced efficiency, accelerated degradation, and hazardous conditions such as thermal runaway. This issue is further exacerbated in high-performance EVs, where elevated power densities and demanding operating conditions lead to larger thermal gradients and greater temperature nonuniformity within the battery system. 

Battery thermal management systems (BTMSs), particularly liquid cooling, have become the standard solution for heat dissipation in EV battery packs \cite{kabirzadeh2025integrating}. Among various thermal management strategies, indirect liquid cooling offers a favorable balance among performance, safety, and cost-effectiveness \cite{tong2025comprehensive}. For pouch cells, indirect liquid cooling typically employs cold plates affixed to the cell surface \cite{gopinadh2022lithium}. These cold plates are engineered with embedded channels that circulate coolant, efficiently dissipating heat from the cells and maintaining a uniform temperature distribution \cite{liu2023data,zheng2022electrical}. However, the effectiveness of BTMSs for pouch cells is primarily dictated by the design of the cooling channels, as these configurations critically influence the uniformity of temperature distribution. 

Accurate estimation of the temperature distribution within pouch cells is therefore critically important for informed thermal management and design optimization. Finite element (FE) analysis is a viable approach for predicting temperature profiles across various configurations, thereby minimizing reliance on extensive experimental testing \cite{qin2022external}. However, the substantial computational demands of high-fidelity simulations limit the practicality of employing such methods for iterative optimization in battery thermal management systems. Consequently, there is a pressing need for surrogate modeling techniques that can provide reliable predictions of temperature distributions with significantly reduced computational overhead while preserving the necessary level of accuracy.
Traditional data-driven methods leverage historical data to predict battery temperature distributions under different configurations. While these approaches can capture complex patterns and provide rapid predictions, their reliability is fundamentally tied to the quantity and diversity of available training data. 
In scenarios where experimental or simulated data are sparse, conventional models often fail to generalize accurately. Furthermore, purely data-driven techniques may overlook critical physical constraints and boundary conditions, sometimes producing implausible results \cite{sutharssan2015prognostic}.

\begin{figure*}[t]
    \centering
    \includegraphics[width=0.7\textwidth]{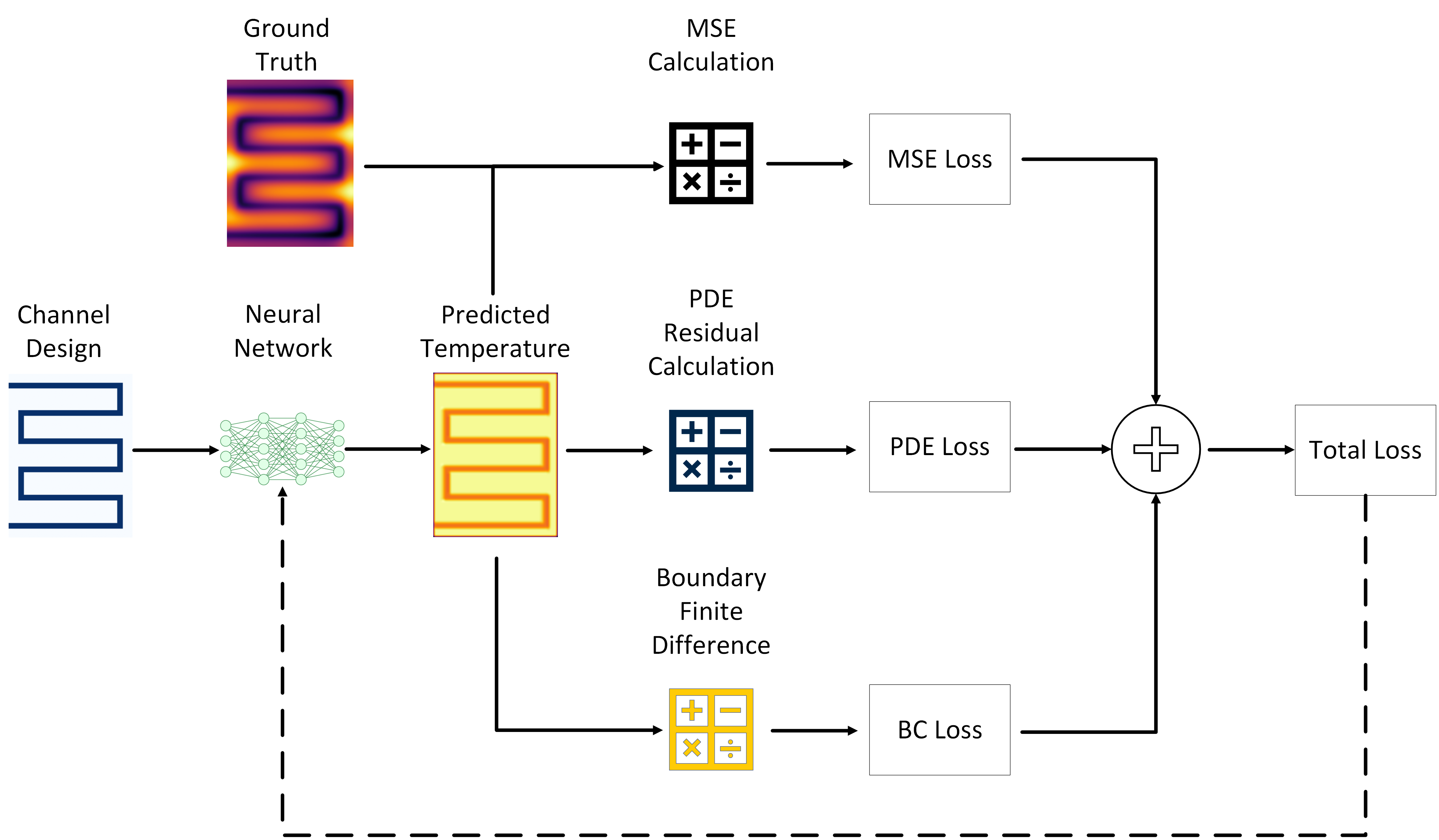}
    \caption{Schematic of the PIML framework for temperature estimation.}
    \label{fig:pinn_arch}
\end{figure*}

Physics-informed machine learning (PIML) integrates physical laws directly into the temperature estimation process, combining data-driven approaches with relevant governing equations \cite{liu2025physics,liu2025physics2,jiang2025multi}. By embedding physical laws into a neural network, PIML can produce predictions consistent with established physical relationships, even when data are limited \cite{karniadakis2021physics}. This integration of physical principles with machine learning substantially improves the model’s reliability, thereby rendering it highly effective for estimating temperature distributions in batteries. By embedding domain knowledge into the modeling process, PIML overcomes the inherent limitations of purely data-driven techniques and provides a robust framework for accurate, computationally efficient thermal analysis in advanced battery management systems.

This research focuses on the application of PIML to accurate temperature estimation in pouch cells with cold plates. Building upon simulations, the simplified two-dimensional heat conduction problem facilitates computational efficiency. Further simplifications of cooling channel configurations and battery heat generation enable derivation of a tractable form of the governing heat transfer equation, which is subsequently embedded in the physics-informed loss function. The temperature distribution across the battery cell is then predicted using a physics-informed convolutional neural network that integrates data-driven insights and physical constraints to enhance estimation accuracy and reliability.

\section{Methods}

The pouch cell chosen for this study has dimensions of 154 $\times$ 203 $\times$ 7.2 mm and can be simplified to a 2D problem \cite{hu2022measurement}. To further simplify the problem, the time-dependent process for battery cycling is converted to a steady-state problem based on the heat generation rate ($Q_{batt}$) generally lower than 8.0 W in most SOCs. First, the Finite Difference Method (FDM) is used to generate temperature distribution data. Then, a data-driven method is developed to accelerate the estimation process. Finally, the PIML framework is established to further enhance the estimation accuracy with less data.

\subsection{Finite Difference Method}

A numerical framework couples a parametric geometry generator with a FDM thermal solver to create a comprehensive dataset for subsequent analysis was established. The 2D area of the pouch cell contact with cold plate is discretized into a structured grid of resolution $N_x \times N_y = 154 \times 203$. Based on the width ($L_x$) and length ($L_y$), this resolution ensures a uniform spatial step size ($dx, dy$) of 1~mm. The steady-state temperature estimation problem can be considered as a 2D heat transfer problem:

To simplify the problem, the temperature of the battery surface is assumed to be equal to that of the thin cold plate. The temperature distribution $T$ of the battery surface is modeled using the steady-state 2D heat diffusion equation for a thin plate with internal heat generation and localized convective cooling. 

\begin{equation}
    k \cdot t \cdot \nabla^2 T + q_{gen} - h(x,y) \cdot (T - T_{coolant}) = 0
\end{equation}
where $k$ is the thermal conductivity of the cold plate, $t$ is the thickness of the cold plate (202.4 $W/(m\cdot K)$\cite{qian2016thermal}), $\nabla^2 T = \frac{\partial^2 T}{\partial x^2} + \frac{\partial^2 T}{\partial y^2}$ is the Laplacian operator representing 2D heat conduction, $q_{gen} = \frac{Q_{batt}}{L_x L_y}$ is the uniform heat generation flux from the battery. $h(x,y)$ is the convection coefficient. $T_{coolant}$ is the coolant temperature.
For this thin (2 mm) cold plate \cite{monika2021numerical}, $h(x,y)$ can be obtained using a binary channel footprint mask $M(x,y)\in\{0,1\}$.
\begin{equation}
h(x,y) = h_{\mathrm{coeff}}\,M(x,y) + h_{\mathrm{bg}}\bigl(1-M(x,y)\bigr),
\label{eq:hmap}
\end{equation}
where $h_{\mathrm{coeff}}= 1500  W/(m^2 \cdot K)$ is the effective convection coefficient in regions under the channels \cite{huo2015investigation} and $h_{\mathrm{bg}}=0$ is a background value.

The in-plane outer edges of the modeled interface are taken as adiabatic.
\begin{equation}
\left.\frac{\partial T}{\partial n}\right|_{\partial\Omega}=0
\quad\text{on}\quad x=\{0,L_x\}\ \text{and}\ y=\{0,L_y\}
\label{eq:neumann}
\end{equation}
where $n$ denotes the outward normal direction.

\subsection{Data-Driven Method}

To accelerate the design optimization process, a deep learning-based surrogate model was developed to approximate the temperature distribution.
The dataset generated by FDM, consisting of $N$ pairs of binary channel masks ($\mathbf{X}$) and corresponding steady-state temperature fields ($\mathbf{T}$), serves as the ground truth for training.
First, the data tensors were reshaped to include a channel dimension, resulting in dimensions $(N, 1, N_y, N_x)$, compatible with 2D convolutional operations. The dataset was randomly partitioned into a training set (80\%) and a testing set (20\%) to evaluate generalization performance.
To ensure numerical stability and efficient convergence during training, the target temperature fields were standardized. The mean ($\mu_{train}$) and standard deviation ($\sigma_{train}$) were computed exclusively from the training dataset:

\begin{equation}
    \mathbf{T}_{norm} = \frac{\mathbf{T} - \mu_{train}}{\sigma_{train}}
\end{equation}

\begin{figure*}[t]
    \centering
    \includegraphics[width=0.6\textwidth]{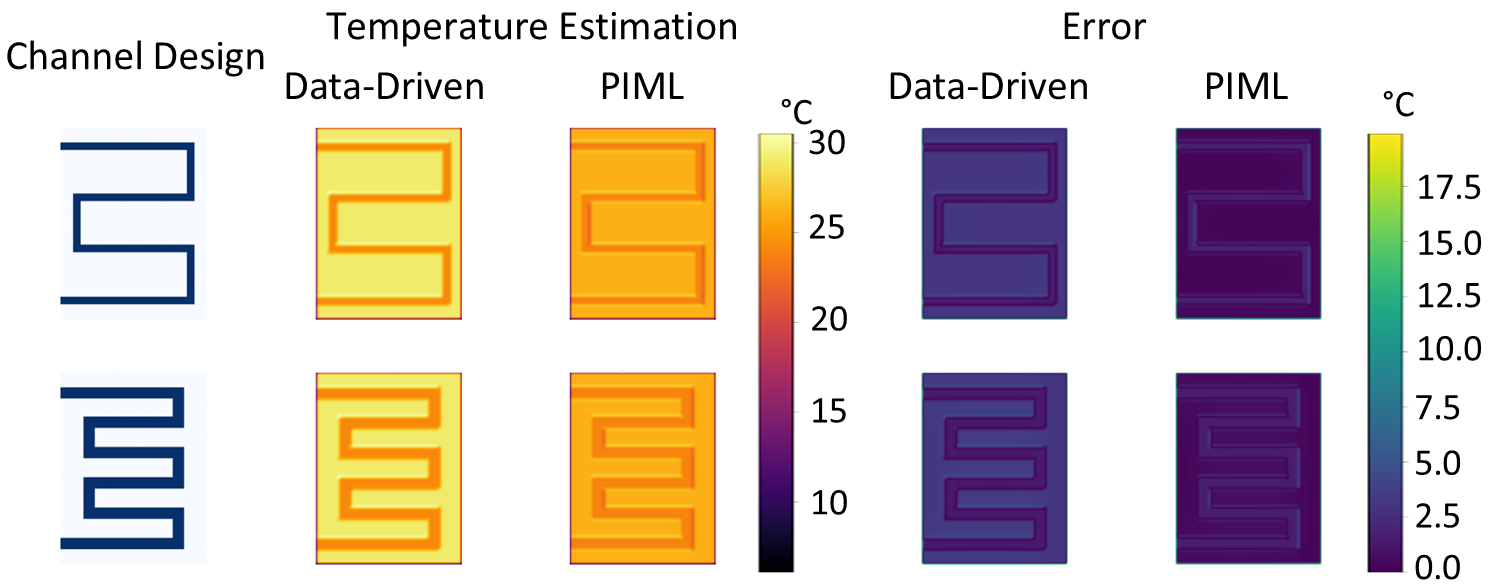}
    \caption{Temperature estimation by data-driven and PIML methods.}
    \label{fig:result}
\end{figure*}

The surrogate model was designed as a Fully Convolutional Network (FCN), preserving spatial resolution throughout the network and mapping the input geometry directly to the output temperature estimate.
Rectified Linear Unit (ReLU) activation functions are applied after the first two layers to introduce non-linearity, allowing the model to capture complex thermal diffusion patterns. The final layer uses a linear activation to predict continuous scalar values representing the normalized temperature.

The training objective was to minimize the Mean Squared Error (MSE) between the predicted normalized temperature maps ($\hat{\mathbf{T}}$) and the ground truth ($\mathbf{T}_{norm}$):

\begin{equation}
    \mathcal{L}_{MSE} = \frac{1}{N_{batch}} \sum_{i=1}^{N_{batch}} || \mathbf{T}_{norm}^{(i)} - \hat{\mathbf{T}}^{(i)} ||^2
\end{equation}

Training was performed for 100 epochs, and the network parameters were optimized using the Adam optimizer with a learning rate of $\alpha = 0.001$. The training process was conducted for 100 epochs. At inference time, the model predictions are inversely transformed using $\mu_{train}$ and $\sigma_{train}$ to recover the temperature values in degrees Celsius.

\subsection{Physics-Informed Machine Learning}

To further improve the temperature estimation, a PIML framework was implemented. Unlike the data-driven surrogate model, the Physics-Informed Neural Network (PINN) directly learns the solution function $T(x,y)$ for a specific geometry by minimizing the residual of the governing partial differential equation (PDE).
The PINN takes continuous spatial coordinates $(x, y)$ as inputs and outputs the predicted temperature scalar $T(x, y)$.

As shown in Fig. \ref{fig:pinn_arch}, the optimization objective is to minimize a composite loss function $\mathcal{L}_{total}$ derived from $\mathcal{L}_{MSE}$ and the physical laws governing the system:

\begin{equation}\label{eq:loss}
    \mathcal{L}_{total} = w_1 \cdot \mathcal{L}_{MSE} + w_2 \cdot \mathcal{L}_{PDE} + w_3 \cdot \mathcal{L}_{BC}
\end{equation}
where $\mathcal{L}_{PDE}$ is the PDE residual loss, and $\mathcal{L}_{BC}$ is the boundary condition loss.  $\mathcal{L}_{PDE}$ is the mean squared error of this residual, normalized by the heat generation term to maintain numerical stability

\begin{equation}
    \mathcal{L}_{PDE} = \frac{1}{N} \sum_{i=1}^{N} \left( \frac{R_i}{q_{\text{gen}}} \right)^2
\end{equation}

The adiabatic boundary conditions are enforced as a soft penalty. The gradients of the temperature field are evaluated at the domain boundaries ($\Omega$). The loss penalizes any non-zero heat flux normal to the boundary.

\begin{equation}
    \mathcal{L}_{BC} = \frac{1}{N} \sum_{j \in \Omega} \left( \left(\frac{\partial T}{\partial x}\right)^2_{j} + \left(\frac{\partial T}{\partial y}\right)^2_{j} \right)
\end{equation}

The spatial domain is discretized into a dense grid of collocation points with the exact resolution as the FDM solver for direct comparison. By minimizing $\mathcal{L}_{total}$, the network converges to a solution.

\section{Results}

Utilizing various computational methods, the temperature distribution within battery pouch cells can be obtained. For the purposes of this study, results derived from the FDM are designated as the ground truth. To generate a comprehensive training dataset, the shape and width of cooling channels are systematically varied, producing 100 distinct samples, each corresponding to a unique channel configuration and its associated temperature distribution.

Both a conventional data-driven model and a PIML model are trained using these 100 samples. Notably, the PIML model demonstrates superior convergence during training: after 10 epochs, the mean squared error (MSE) loss for the PIML model declines to 5.66, compared with 11.12 for the data-driven model, representing a 49.1\% improvement in performance.

A separate validation dataset is employed to rigorously assess both models. As illustrated in Fig. \ref{fig:result}, various channel designs are evaluated based on the resulting temperature distributions. The PIML model consistently provides more accurate temperature estimations than the purely data-driven approach, particularly in regions outside the immediate path of the cooling channels. These findings underscore the advantages of PIML in modeling the thermal behavior of battery pouch cells equipped with cold plates, offering enhanced accuracy for practical thermal management applications.

\section{Conclusion and Future Work}
This study demonstrates that PIML provides significant improvements in temperature estimation accuracy for pouch cells with indirect liquid cooling, compared to purely data-driven approaches. By incorporating governing heat transfer equations into the training process, PIML achieves faster convergence and enhanced reliability, particularly in regions beyond the cooling channels. These results highlight the potential of PIML as an efficient surrogate modeling tool for BTMS design, thereby reducing both experimental costs and reliance on extensive simulation data.
Future research will extend the PIML framework to optimize advanced cooling channel designs, complemented by rigorous experimental validation. This will enable a more comprehensive assessment of the model’s predictive capabilities and further support practical implementation in real-world BTMSs.

\bibliographystyle{ieeetr}
\bibliography{ref}

\end{document}